\documentclass[letterpaper]{article} 
\usepackage[submission]{aaai23}  
\usepackage{times}  
\usepackage{helvet}  
\usepackage{courier}  
\usepackage[hyphens]{url}  
\usepackage{graphicx} 
\usepackage{natbib}  
\usepackage{caption} 
\usepackage{algorithm}
\usepackage{algorithmic}
\usepackage{amsmath}
\usepackage{amsfonts}

\DeclareMathOperator*{\argmin}{arg\,min}

\usepackage{newfloat}
\usepackage{listings}
\DeclareCaptionStyle{ruled}{labelfont=normalfont,labelsep=colon,strut=off} 
\floatstyle{ruled}
\newfloat{listing}{tb}{lst}{}
\floatname{listing}{Listing}
\title{Hierarchical Needs-driven Agent Learning Systems: \\ From Deep Reinforcement Learning To Diverse Strategies}
\author{
    Qin Yang
}
\affiliations{
    The Department of Computer Science at the University of Georgia \\
    Email: RickYang2014@gmail.com; qy03103@uga.edu

}

\usepackage{bibentry}

\begin{document}

\maketitle

\begin{abstract}
The needs describe the necessities for a self-organizing system to survive and evolve, which arouses an agent to action toward a goal, giving purpose and direction to behavior. Based on Maslow’s hierarchy of needs, an agent needs to satisfy a certain amount of needs at the current level as a condition to arise at the next stage -- upgrade and evolution. Especially, Deep Reinforcement Learning (DAL) can help AI agents (like robots) organize and optimize their behaviors and strategies to develop diverse Strategies based on their current state and needs (expected utilities or rewards). This paper introduces the new hierarchical needs-driven Learning systems based on DAL and investigates the implementation in the single-robot with a novel approach termed Bayesian Soft Actor-Critic (BSAC). Then, we extend this topic to the Multi-Agent systems (MAS), discussing the potential research fields and directions.

\end{abstract}

\section{Introduction}
In nature, the interaction between and within various elements of a system is complex \cite{chan2001complex}. Many natural systems (e.g., brains, immune system, ecology, societies) and artificial systems (robotic and AI systems) are characterized by apparently complex behaviors that emerge as a result of often nonlinear spatiotemporal interactions among a large number of component systems at different levels of the organization \cite{levin1998ecosystems}. For example, to maximize the chance of detecting predators, forage, and save more energy while migrating to different locations, animals usually split into different swarms to minimize their aggregated threat and maximize benefits according to different scenarios forming complex adaptive systems. Formally, according to the {\it expected utility hypothesis} \cite{von2007theory}, considering the probabilities of agents' needs in the specific situation, their decisions always want to maximize expected utilities (needs) and also maximize the probability of the decision's consequences being preferable to some uncertain threshold. Furthermore, as the combination of hierarchical needs, intelligent agents develop various strategies and skills to adapt to different scenarios, satisfying or maximizing their current dominant needs based on the corresponding rewards (expected utility/needs) mechanism through learning from the interaction with other agents and environments.

\begin{figure}[tbp]
\centering
\includegraphics[width=1.04\columnwidth]{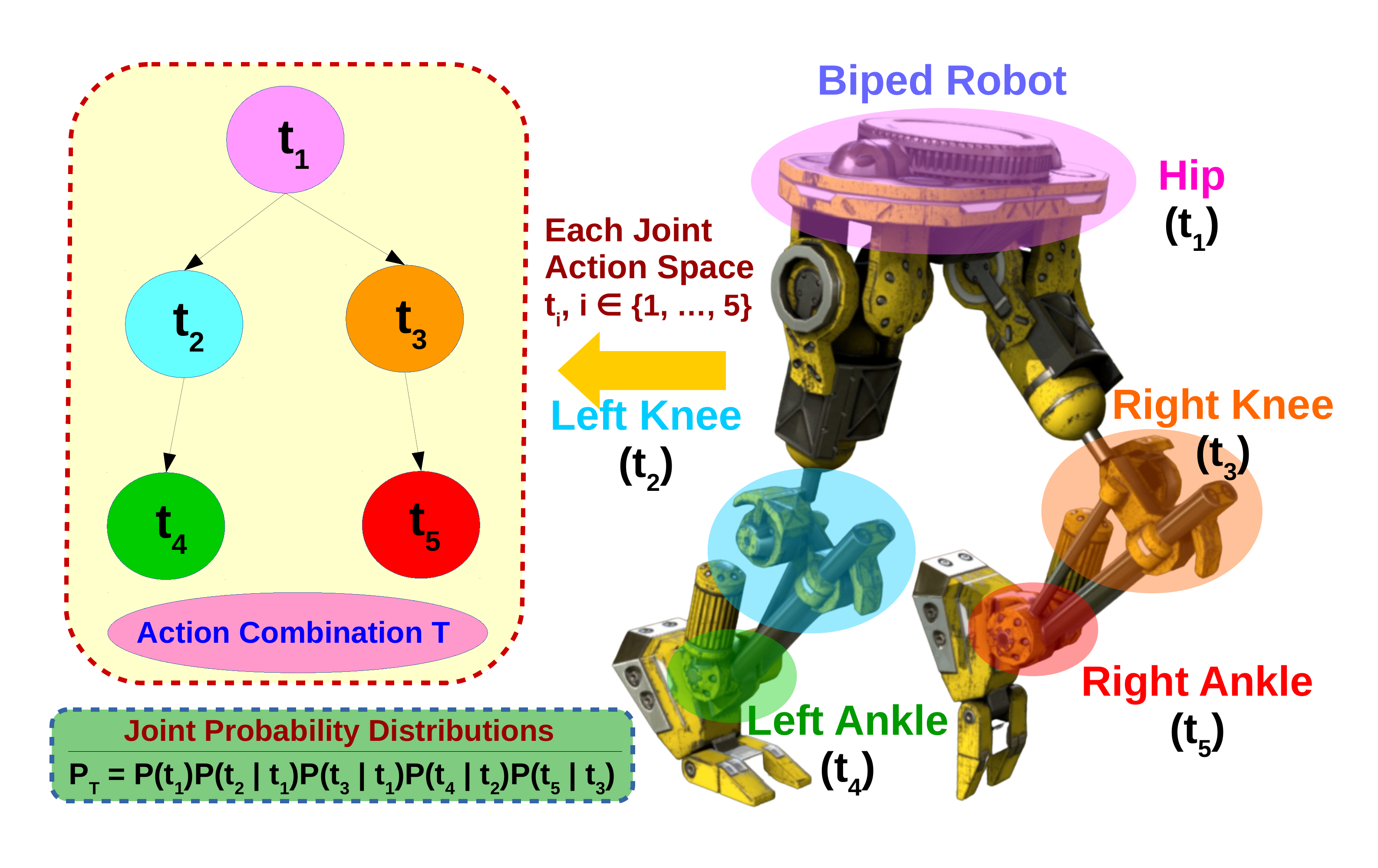}
\caption{An example biped robot model's Bayesian Strategy Network, showing a decomposed strategy based on action dependencies.}
\label{fig:overview}
\end{figure}

In Artificial Intelligence (AI) methods, a strategy describes the general plan of an AI agent\footnote{This paper uses the terms agent and robot interchangeably.} achieving short-term or long-term goals under conditions of uncertainty, which involves setting sub-goals and priorities, determining action sequences to fulfill the tasks, and mobilizing resources to execute the actions \cite{freedman2015strategy}. It exhibits the fundamental properties of agents' perception, reasoning, planning, decision-making, learning, problem-solving, and communication in interaction with dynamic and complex environments \cite{langley2009cognitive,yang2019self}. Especially in the field of real-time strategy (RTS) game \cite{buro2003real,yang2022game} and real-world implementation scenarios like robot-aided urban search and rescue (USAR) missions \cite{murphy2014disaster,yang2020needs}, agents need to dynamically change the strategies adapting to the current situations based on the environments and their expected utilities or needs \cite{yang2020hierarchical, yang2021can}.

\begin{figure*}[tbp]
\centering
\includegraphics[width=2.1\columnwidth]{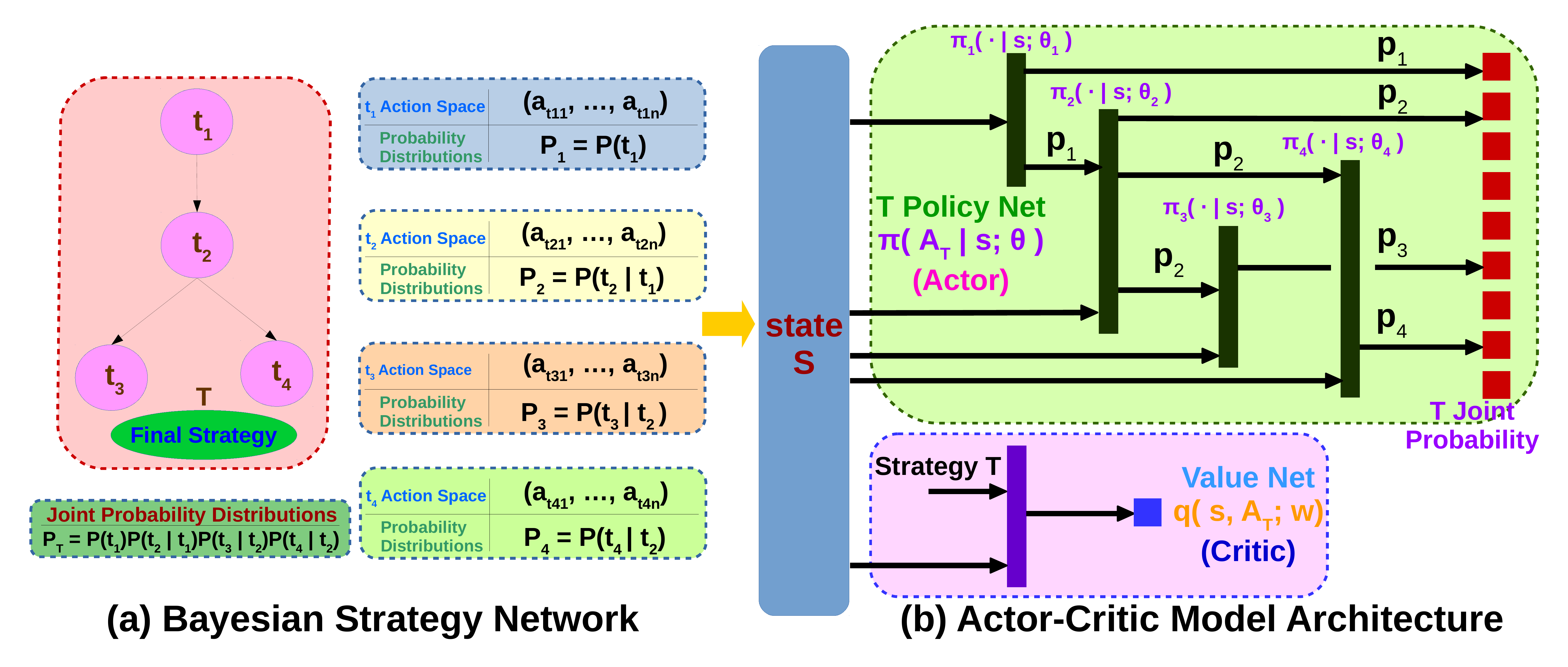}
\caption{\normalsize An overview of the proposed BSN based implementation of the Actor-Critic DRL architecture model}
\label{bsn_drl}
\end{figure*}

From the agent perspective, a strategy is a rule used by agents to select an action to pursue goals, which is equivalent to a policy in a Markov Decision Process (MDP) \cite{rizk2018decision}. In reinforcement learning (RL), the policy dictates the actions that the agent takes as a function of its state and the environment, and the goal of the agent is to learn a policy maximizing the expected cumulative rewards in the process. With advancements in deep neural network implementations, deep reinforcement learning (DRL) helps AI agents master more complex strategies (policies) and represents a step toward building autonomous systems with a higher-level understanding of the visual world \cite{arulkumaran2017deep}. 

Specifically, reinforcement learning is a framework that helps develop self-learning capability in robots, but it is limited to the lower-dimensional problem because of complexity in memory and computation; Deep RL integrates the deep neural network implementing {\it function approximation} and {\it representation learning} to overcome the limitation of RL \cite{singh2021reinforcement}. On the other hand, current research and industrial communities have sought more software-based control solutions using low-cost sensors with less operating environment requirements and calibration \cite{liu2021deep}. As the most promising algorithm, DRL ideally suits robotic manipulation and locomotion because of no predefined training data requirement. Furthermore, the control policy could be obtained by learning and updating instead of hard-coding directions to coordinate all the joints.

\begin{figure*}[t]
\centering
\begin{minipage}[b]{0.265\linewidth}
\includegraphics[width=1\textwidth]{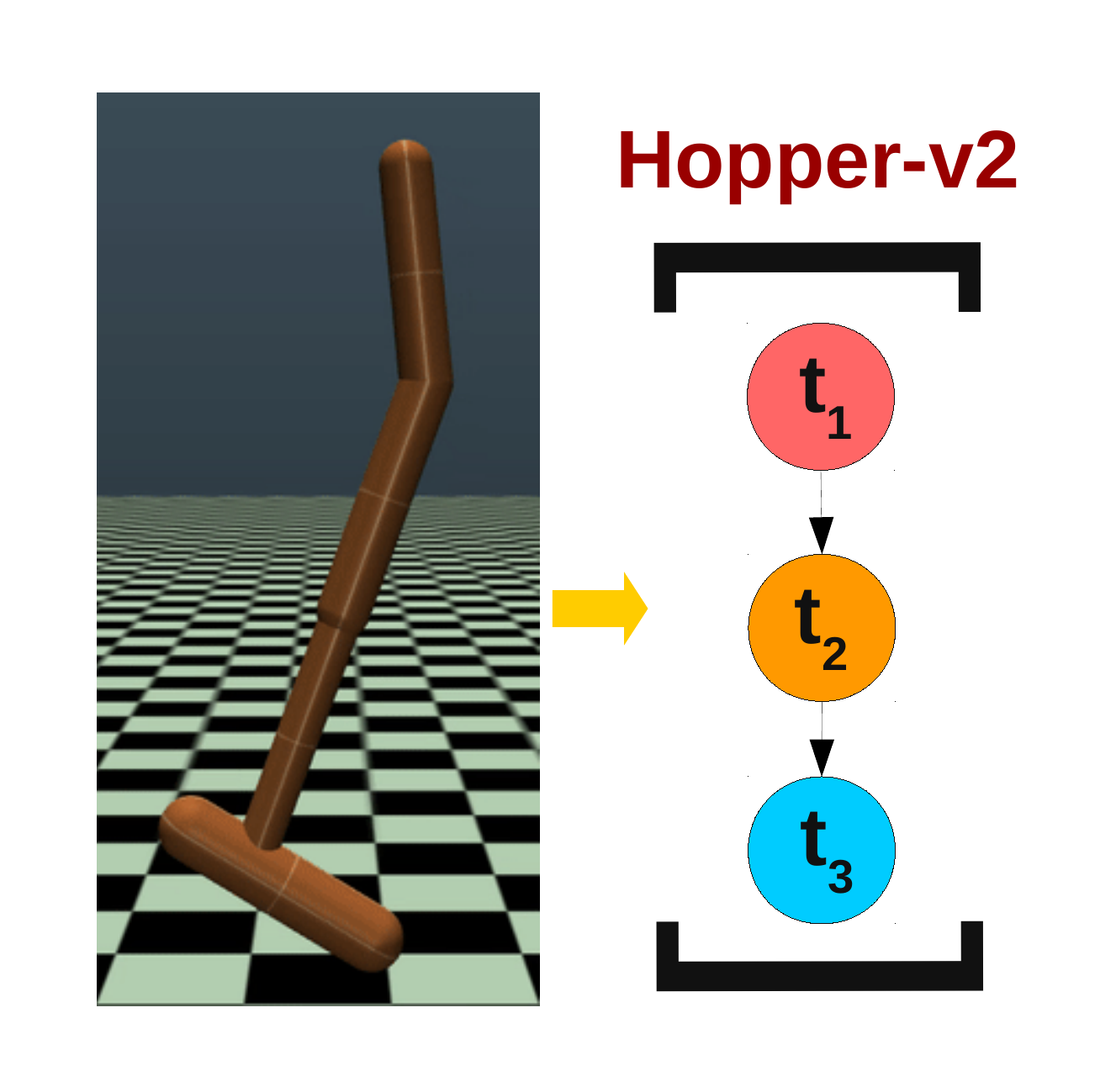}\vspace{0pt}
\caption{\small{BSN 3P Model in Hopper}}
\label{hopper_2v_bsn}
\end{minipage}
\begin{minipage}[b]{0.314\linewidth}
\includegraphics[width=1\textwidth]{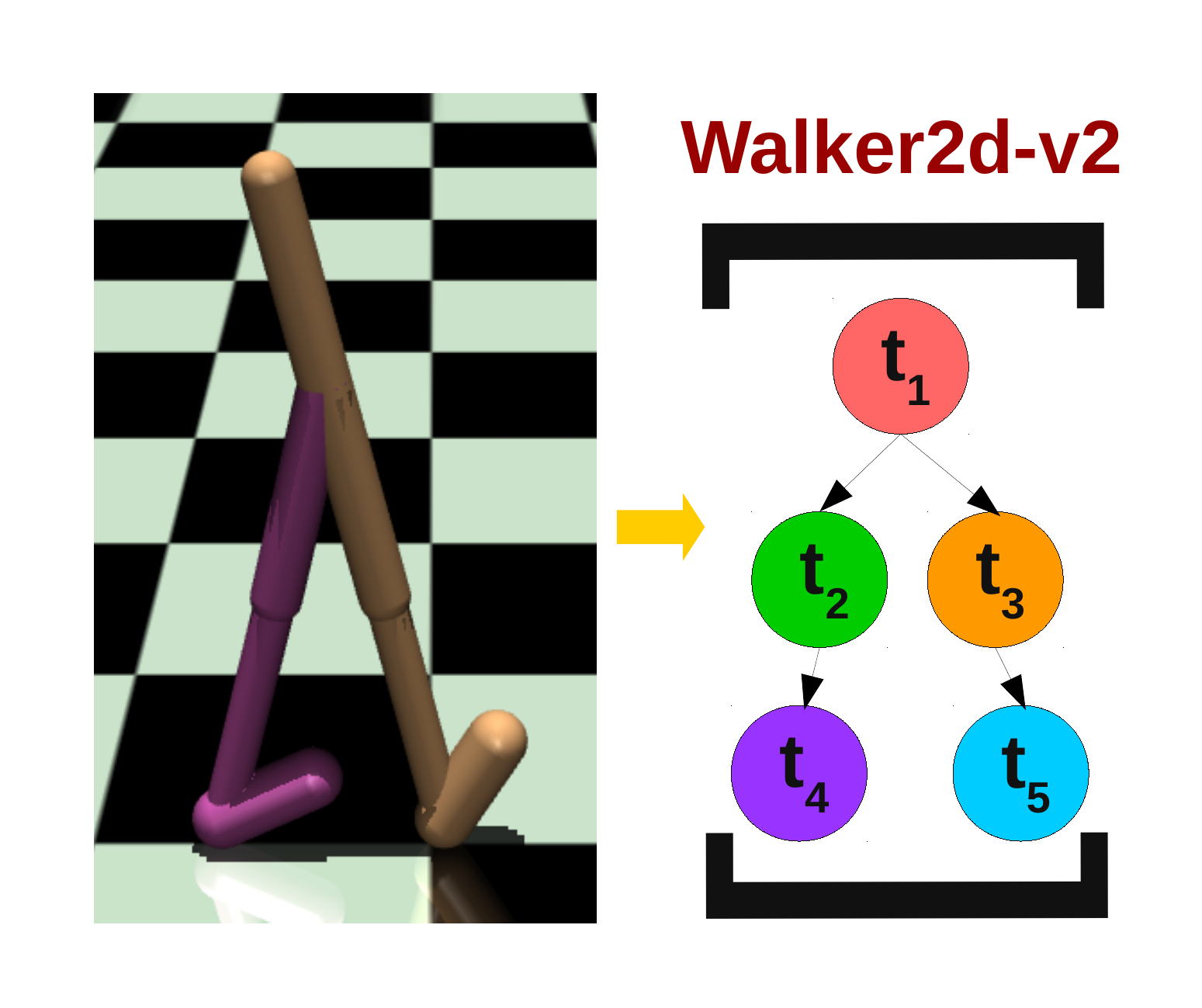}\vspace{0pt}
\caption{\small{BSN 5P Model in Walker2d}}
\label{walker2d_2v_bsn}
\end{minipage}
\begin{minipage}[b]{0.38\linewidth}
\includegraphics[width=1\textwidth]{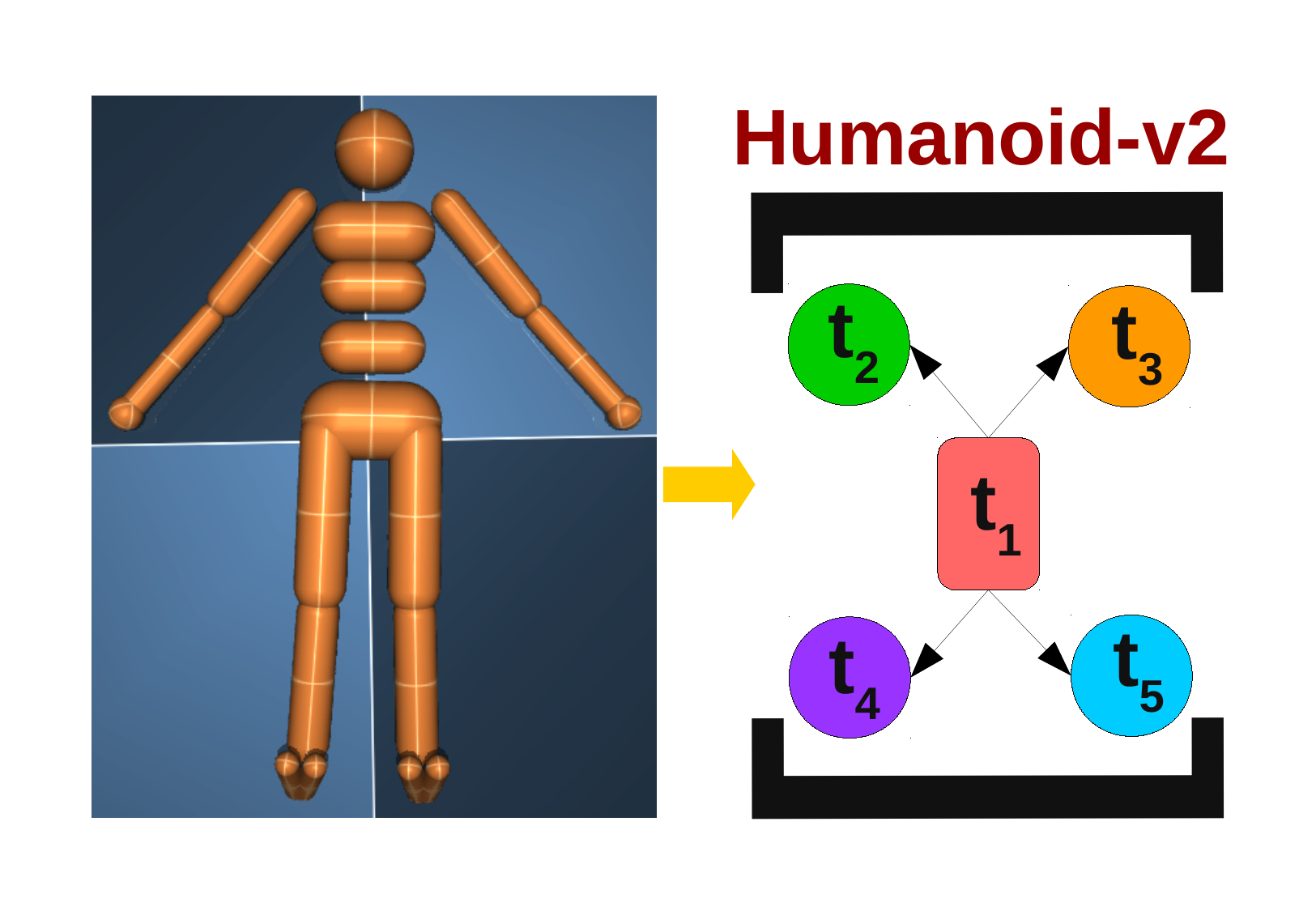}\vspace{0pt}
\caption{\small{BSN 5P Model in Humanoid}}
\label{fig:humanoid_5bsn}
\end{minipage}
 \vspace{-1mm}
\end{figure*}

\section{Background and Preliminaries}

This section provides the essential background about \textit{Agent Needs Hierarchy} and \textit{Deep Reinforcement Learning} When describing a specific method, we use the notations and relative definitions from the corresponding papers.

\subsection{Agent Needs Hierarchy}
In \textit{Agent Needs Hierarchy} \cite{yang2020hierarchical}, the abstract needs of an agent for a given task are prioritized and distributed into multiple levels, each of them preconditioned on their lower levels. At each level, it expresses the needs as an expectation over the corresponding factors/features' distribution to the specific group \cite{yang2020needs}. 

Specifically, it defines five different levels of agent needs similar to Maslow's human needs pyramid \cite{yang2021can,yang2020game}. The lowest (first) level is the safety features of the agent (e.g., features such as collision detection, fault detection, etc., that assure safety to the agent, human, and other friendly agents in the environment). The safety needs (Eq.~\eqref{safety_need}) can be calculated through its innate value and the probability of corresponding safety features based on the agent's current state. After satisfying safety needs, the agent considers its basic needs (Eq.~\eqref{basic_need}), which includes features such as energy levels and data communication levels that help maintain the basic operations of that agent. 
Only after fitting the safety and basic needs, an agent can consider its capability needs (Eq.~\eqref{capability_need}), which are composed of features such as its health level, computing (e.g., storage, performance), physical functionalities (e.g., resources, manipulation), etc. 
At the next higher level, the agent can identify its teaming needs (Eq.~\eqref{teaming_need}) that account for the contributions of this agent to its team through several factors (e.g., heterogeneity, trust, actions) that the team needs so that they can form a reliable and robust team to perform a given mission successfully. 

Ultimately, at the highest level, the agent learns skills/features to improve its capabilities and performance in achieving a given task, such as Reinforcement Learning.
The policy features (Q table or reward function) are accounted into its learning needs expectation (Eq.~\eqref{learning_need}).
The expectation of agent needs at each level are given below:
\vspace{-2mm}
\begin{equation}
\begin{split}
    Safety~~Needs: N_{s_{j}} = \sum_{i=1}^{s_{j}} S_{i} \mathbb{P}(S_{i}|X_j, T); \label{safety_need}
\end{split}
\end{equation}
\vspace{-2mm}
\begin{equation}
\begin{split}
    Basic~~Needs: N_{b_j} = \sum_{i=1}^{b_{j}} B_{i} \mathbb{P}(B_{i}|X_j, T, N_{s_{j}}); \label{basic_need}
\end{split}
\end{equation}
\vspace{-2mm}
\begin{equation}
\begin{split}
    Capability~Needs: N_{c_j} = \sum_{i=1}^{c_{j}} C_i \mathbb{P}(C_i|X_j, T, N_{b_j}); \label{capability_need}
\end{split}
\end{equation}
\vspace{-2mm}
\begin{equation}
\begin{split}
    Teaming~~Needs: N_{t_j} = \sum_{i=1}^{t_j} T_i \mathbb{P(}T_i | X_j, T, N_{c_j}); \label{teaming_need}
\end{split}
\end{equation}
\vspace{-2mm}
\begin{equation}
\begin{split}
    Learning~~Needs: N_{l_j} = \sum_{i=1}^{l_j} L_i \mathbb{P(}L_i | X_j, T, N_{t_j}); \label{learning_need}
\end{split}
\end{equation}
Here, $X_j=\{P_j,C_j\}$ $\in$ $\Psi$ is the combined state of the agent $j$ with $P_j$ being the perceived information by that agent and $C_j$ representing the communicated data from other agents. $T$ is the assigned task. $S_i$, $B_i$, $C_i$, $T_i$, and $L_i$ are the utility values of corresponding feature/factor $i$ in the corresponding levels - Safety, Basic, Capability, Teaming, and Learning, respectively. 
$s_j$, $b_j$, $c_j$, $t_j$, and $l_j$ are the sizes of agent $j$'s feature space on the respective levels of needs.
The collective need of an agent $j$ is expressed as the union of needs at all the levels in the needs hierarchy as in Eq.~\eqref{eqn:need-union} \footnote{Each category needs level is combined with various similar needs (expected values) presenting as a set, consisting of individual hierarchical and compound needs matrix $N_j$.}.
\begin{equation}
    N_j = N_{s_j} \cup N_{b_j} \cup N_{c_j} \cup N_{t_j} \cup N_{l_j} 
    \label{eqn:need-union}
\end{equation}

More specifically, the set of agent needs in a multi-agent system can be regarded as a kind of motivation or requirements for cooperation between agents to achieve a specific group-level task \cite{yang2022self}.

\subsection{Deep Reinforcement Learning (DRL)}

The essence of reinforcement learning (RL) is learning from interaction based on reward-driven (like utilities and needs) behaviors, much like natural agents. When an RL agent interacts with the environment, it can observe the consequence of its actions and learn to change its behaviors based on the corresponding rewards received. Moreover, the theoretical foundation of RL is the paradigm of trial-and-error learning rooted in behaviorist psychology \cite{sutton2018reinforcement}. Furthermore, DRL trains deep neural networks to approximate the optimal policy and/or the value function. The deep neural network serving as a function approximator enables powerful generalization, especially in visual domains, general AI systems, robotics, and multiagent/robot systems (MAS/MRS) \cite{hernandez2019survey}. The various DRL methods can be divided into three groups: value-based methods, such as DQN \cite{mnih2015human}; policy gradient methods, like the PPO \cite{schulman2017proximal}; and actor-critic methods, like the Asynchronous Advantage Actor-Critic (A3C) \cite{mnih2016asynchronous}. From the deterministic policy perspective, DDPG \cite{lillicrap2015continuous} provides a sample-efficient learning approach. On the other hand, from the entropy angle, SAC \cite{haarnoja2018soft} considers a more general maximum entropy objective retaining the benefits of efficiency and stability. Here, we briefly discuss SAC method as follows:


\paragraph{Soft Actor-Critic (SAC):} SAC is an off-policy actor-critic algorithm that can be derived from a maximum entropy variant of the policy iteration method. The architecture consider a parameterized state value function $V_\psi(s_t)$, soft Q-function $Q_\theta(s_t, a_t)$, and a tractable policy $\pi_\phi(a_t | s_t)$. It updates the policy parameters by minimizing the Kullback-Leibler divergence between the policy $\pi'$ and the Boltzmann policy in Eq. \eqref{sac_pi}.
\begin{equation}
\begin{split}
    \pi_{new} = \argmin_{\pi'} D_{KL} \left( \pi'(\cdot | s_t) \bigg\Vert \frac{exp(Q_\theta(s_t, \cdot)}{Z_\theta(s_t)} \right)
\label{sac_pi}
\end{split}
\end{equation}


\begin{figure*}[t]
\centering
\includegraphics[width=2.1\columnwidth]{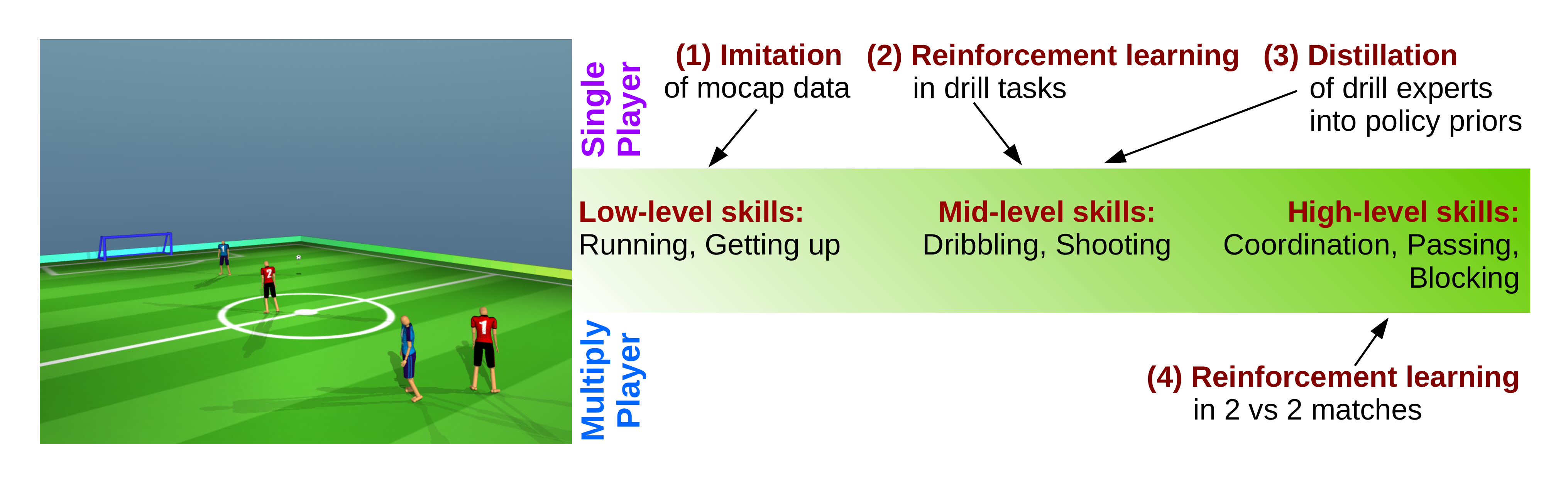}
\caption{\normalsize End-to-End learning of coordinated 2 vs 2 humanoid football in MuJoCo multi-agent soccer environment}
\label{fig:humanoid_football}
\end{figure*}

\section{Hierarchical Needs-driven Agent learning}

This section first introduces the Bayesian Strategy Network (BSN) \cite{yang2022bsac} based on the Bayesian networks to decompose a complex strategy or intricate behavior into several simple tactics or actions. Then, from a single-agent perspective, we discuss the promising potential of a new DRL model termed Bayesian Soft Actor-Critic (BSAC) in robotics. An example of a BSN-based action decomposition of a Biped Robot is shown in Fig.~\ref{fig:overview}.
Furthermore, we extend this topic to MAS/MRS fields.

\subsection{Single-Agent Systems}
\paragraph{Bayesian Strategy Networks (BSN)}
Supposing that the strategy $\mathcal{T}$ consists of $m$ tactics ($t_1, \dots, t_m$) and their specific relationships can be described as the BSN. We consider the probability distribution $P_{i}$ as the policy for tactic $t_i$. Then, according to the Bayesian chain rule, the joint policy $\pi(a_{\mathcal{T}} \in {\mathcal{T}},s)$ can be described as the joint probability function (Eq. \eqref{agent_task_policy}, $m \in Z^+$) through each sub-policy $\pi_i(t_i,s)$, correspondingly. An overview of the example BSN implementation in actor-critic architecture is presented in Fig. \ref{bsn_drl}.
\begin{equation}
\begin{split}
    \pi_{\mathcal{T}}(t_1, \dots, t_m) = \pi_{1}(t_1) \prod_{i=2}^{m} \pi_{i}(t_i | t_1, \dots, t_i).
\label{agent_task_policy}
\end{split}
\end{equation}

Fig.~\ref{hopper_2v_bsn}, \ref{walker2d_2v_bsn}, and \ref{fig:humanoid_5bsn} provide three examples of decomposing RL agents' behaviors into different BSN models in the standard continuous control benchmark domains -- Hopper-v2, Walker2d-v2, and Humanoid-v2 -- at OpenAI's Gym MuJoCo environments. The source of the BSAC algorithm is available on GitHub: \url{https://github.com/RickYang2016/Bayesian-Soft-Actor-Critic--BSAC}.

\paragraph{Bayesian Soft Actor-Critic (BSAC)}
BSAC integrates the Bayesian Strategy Networks (BSN) and the state-of-the-art SAC method \cite{yang2022bsac}. By building several simple sub-policies organized as BSN, BSAC provides a more flexible and suitable joint policy distribution to adapt to the Q-value distribution, increasing sample efficiency and boosting training performance.

Specifically, BSAC incorporates the maximum entropy concept into the actor-critic deep RL algorithm. According to the additivity of the entropy, the system's entropy can present as the sum of the entropy of several independent sub-systems \cite{wehrl1978general}. For each step of soft policy iteration, the joint policy $\pi$ will calculate the value to maximize the sum of sub-systems' $\pi_i$ entropy in the BSN using the below objective function Eq. \eqref{bsac_entropy}. To simplify the problem, each policy's weight and corresponding temperature parameters $\alpha_i$ are the same in each sub-system.
\begin{equation}
\begin{split}
    J_V(\pi) = \sum_{t=0}^{T} \mathbb{E}_{(s_t, \mathcal{A}_t) \sim \rho_{\pi_i}} \left [r(s_t, \mathcal{A}_t) + \frac{\alpha}{m} \sum_{i=1}^{m} \mathcal{H}(\pi_i(\cdot|s_t)) \right ]
\label{bsac_entropy}
\end{split}
\end{equation}

Furthermore, the soft policy evaluation and the soft policy improvement alternating execution in each soft sub-policy iteration guarantees the convergence of the optimal maximum entropy among the sub-policies combination. 

\subsection{Multi-Agent Systems (MAS)}



When considering multi-agent systems (MAS) learning, an individual agent must first master essential skills to satisfy low-level needs (safety, basic, and capability needs). Then, it will develop effective strategies fitting middle-level needs (like teaming and collaboration) to guarantee the systems' utilities. Through learning from interaction, MAS can optimize group behaviors and present complex strategies adapting to various scenarios and achieving the highest-level needs, and fulfilling evolution. By cooperating to achieve a specific task, gaining {\it expected needs (rewards)}, or against the adversaries decreasing the threat, intelligent agents can benefit the entire group development or utilities and guarantee individual needs and interests.
It is worth mentioning that \cite{liu2021motor} developed the end-to-end learning implemented in the MuJoCo multi-agent soccer environment \cite{tunyasuvunakool2020dm_control}, which combines low-level imitation learning, mid-level, and high-level reinforcement learning, using transferable representations of behavior for decision-making at different levels of abstraction. Figure \ref{fig:humanoid_football} represents the training process.

Specifically, {\it Agent Needs Hierarchy} is the foundation for a MAS learning from interaction. It surveys the system’s utility from individual needs. Balancing the needs/rewards between agents and groups for MAS through interaction and adaptation in cooperation optimizes the system’s utility and guarantees sustainable development for each group member, much like human society does. Furthermore, by designing suitable reward mechanisms and developing efficient DRL methods, MAS can effectively represent various group behaviors, skills, and strategies to adapt to uncertain, adversarial, and dynamically changing environments.

\section{Conclusions}

This paper introduces the new hierarchical needs-driven agent learning systems, discusses the implementation of the BSAC in single-agent systems, and extends this topic to the potential research fields in the MAS. It presents the promising potential that hierarchical needs-driven learning systems can bridge the future gap between robotics and AI.

\clearpage

\bibliography{aaai23}

\end{document}